\documentclass{article} 
\usepackage{iclr2025_conference,times}
\iclrfinalcopy
\usepackage{graphicx}
\usepackage{nicefrac}
\usepackage{tikz}
\usetikzlibrary{positioning, arrows, shapes.geometric}

\usepackage{amsmath,amsfonts,bm}

\def\eqref#1{equation~\ref{#1}}

\def\1{\bm{1}}

\DeclareMathAlphabet{\mathsfit}{\encodingdefault}{\sfdefault}{m}{sl}
\SetMathAlphabet{\mathsfit}{bold}{\encodingdefault}{\sfdefault}{bx}{n}

\DeclareMathOperator*{\argmin}{arg\,min}

\usepackage{hyperref}
\usepackage{url}
\usepackage{cleveref}

\title{Fusion of Graph Neural Networks via Optimal Transport}

\author{{\bf Weronika Ormaniec}\thanks{Equal contribution. Correspondence to: wormaniec@ethz.ch, michavol@ethz.ch, elisa.hoskovec@alumni.ethz.ch. Our code is publically available at \href{https://github.com/michavol/gnn_fusion}{https://github.com/michavol/gnn\_fusion}.} \\
{\bf Michael Vollenweider}$^*$ \\
{\bf Elisa Hoskovec}$^*$ \\
Department of Computer Science, ETH Zürich
}

\begin{document}

\maketitle

\vspace{-0.2cm}
\begin{abstract}
In this paper, we explore the idea of combining GCNs into one model. To that end, we align the weights of different models layer-wise using optimal transport (OT). We present and evaluate three types of transportation costs and show that the studied fusion method consistently outperforms the performance of vanilla averaging. Finally, we present results suggesting that model fusion using OT is harder in the case of GCNs than MLPs and that incorporating the graph structure into the process does not improve the performance of the method.
\end{abstract}

\section{Introduction}
\label{sec:introduction}
\vspace{-0.1cm}
In recent years, combining multiple models into one through model fusion has emerged as a new research topic. One of its main uses is the creation of efficient alternatives to ensemble methods that rely on expensive inference \citep{singh2023model,pmlr-v162-wortsman22a,matena2022merging,ainsworth2023git,nguyen2023crosslayer,kandpal2023gittheta,imfeld2023transformer}. \cite{singh2023model} introduced the idea of combining neural networks by aligning neurons layer-wise using optimal transport (OT) and then averaging their weights. They were able to successfully fuse MLPs, CNNs, and RNNs. \cite{imfeld2023transformer} applied a similar approach to Transformers. Finally, \cite{jing2024deep} explored the fusion of GNNs via deep graph mating (see \cref{app:grama}).

In this work, we (1) successfully adapt the OT fusion approach to graph convolutional networks (GCNs), and show preliminary results suggesting that (2) GCNs are harder to fuse with OT than MLPs and (3) incorporating the graph structure into the fusion process is not beneficial.
\section{Background}
\vspace{-0.1cm}
\paragraph{Optimal Transport} \label{sec:sinkhorn}
One of the main ingredients for the proposed approach is optimal transport (OT). OT addresses the problem of optimally transporting a given source distribution to a target distribution. The theory we outline here is strongly inspired by a comprehensive overview by \cite{thorpe}. We use variations of the problem statement for pairs of discrete probability measures $\mu = \sum_{i=1}^{m}\alpha_i\delta_{x_i}$ and $\nu = \sum_{j=1}^{n}\beta_j\delta_{y_j}$, where $\delta_x$ represents the unit mass of some point and $\{x_1,...,x_m\}$ and $\{y_1,...,y_n\}$ are the support of the source and target defined in some shared space $S$. Further, we have that $\alpha$ and $\beta$ live in the $m$ and $n$ dimensional probability simplex, respectively.

\paragraph{EMD Algorithm}
The EMD algorithm solves the Kantorovich formulation of OT given as the linear program
\begin{equation*}\textstyle
    T^* = \argmin_{T \in U(\alpha, \beta)} \sum_{i=1}^m \sum_{j=1}^n C_{ij} T_{ij} = \langle T, C \rangle,
    \vspace{-5pt}
\end{equation*}
where $C\in \mathbb{R}^{m,n}_+$ is the cost matrix with $C_{ij}$ being the cost of transporting point $x_i$ to $y_j$ and $T^*$ the optimal transport matrix (TM) living in $U(\alpha, \beta) := \{T \in \mathbb{R}_+^{m,n} | T\mathbf{1}_n=\alpha, T^T\mathbf{1}_m=\beta$\}. The objective evaluated at $T^*$ is sometimes called the Wasserstein distance between the two distributions. 

\paragraph{Sinkhorn Algorithm}
We require two variations of the Sinkhorn algorithm. The first version solves an entropy-regularized and unbalanced version of the classical linear Kantovorich formulation, allowing for a smoother solution space and some slackness in the balancedness constraint originally imposed on the TM by $U(\alpha, \beta)$. The according objective is given by
\begin{equation*}
    \min_{T \in \mathbb{R}^{n, m}_{+}} \langle T, C \rangle + \rho_{\alpha} KL(T \mathbf{1}_m || \alpha) + \rho_{\beta} KL(T^T \mathbf{1}_n || \beta) - \epsilon KL(T,\alpha \beta^T),
\end{equation*}
where $\epsilon$ controls the regularization strength, parameters $\rho_{\alpha}$, $\rho_{\beta}$ scale the penalization for deviations from full balancedness and $KL(\cdot||\cdot)$ denotes the Kullbeck-Leibler divergence between two distributions. The second version solves a formulation of OT between two probability measures defined on graphs with feature vectors assigned to their vertices. The minimum of the corresponding objective is termed fused Gromov-Wasserstein distance and measure the distance between structured data with feature information.  For a theoretical introduction, we refer the reader to \cite{vayer2019optimal}.

\paragraph{OT Fusion}
\cite{singh2023model} utilizes OT to fuse networks layer by layer, using either weights or activations as support of the underlying distributions. Formally, for a layer $l$, the TM of the previous layer is post-multiplied with the weight matrix of the current layer: $\hat{W}^{(l, l-1)}:= W^{(l, l-1)}T^{(l-1)}\text{diag}(\frac 1 {\beta^{(l-1)}})$. 
The current layer is then aligned by pre-multiplying with the transpose of the TM of the current layer: 
$\tilde{W}^{(l, l-1)} :=\text{diag}(\frac 1 {\beta^{(l)}})T^{(l)^T}\hat{W}^{(l, l-1)}$.
After a network is aligned to an anchor model, their weights are averaged. In this work, we assume that $\beta_i = \beta_j \; \forall{i, j}$.

\section{Models and Methods}
\label{sec:method}
\subsection{Graph Convolutional Networks}

A GCN is a type of neural network designed to operate on graph-structured data. Let  $G = (V, E)$ represent a graph, where $V$ is the set of vertices and $E$ the set of edges. The vertices can have feature vectors associated with them.

The key idea of GCNs is to perform convolutional operations on the graph data, similar to how convolutional layers operate on regular grids in image data. The propagation rule for a single graph convolutional (GC) layer, as introduced by \cite{gcn}, can be expressed as
\begin{equation*}\textstyle
    h^{(l+1)}_i=\text{ReLU}(W^{(l+1,l)}\frac 1 {\sqrt{\text{deg}_i}}\sum_{j \in \mathcal{N}_i}\frac 1 {\sqrt{\text{deg}_j}}h^{(l)}_j + b^{(l+1, l)}).
    \vspace{-5pt}
\end{equation*}
Here, $h^{(l)}_i$ is a feature vector of the $i^{th}$ vertex at layer $l$, $\mathcal{N}_i$ the set of neighbors of vertex $i$ (including vertex $i$ itself), and $W^{(l, l+1)}$ and $b^{(l, l+1)}$ the weights and biases for that layer. In the context of this work, it is important to note that the weight matrix is shared among vertices. Prior to applying the rectified linear unit (ReLU) activation function, one can incorporate batch normalization (BN). 

In this work we consider the GCN architecture proposed in \cite{dwivedi2020benchmarkgnns}. Firstly, vertex features are processed with an embedding layer, then $g$ GC layers with or without BN depending on the experiment are applied sequentially producing a hidden representation of each graph vertex feature. Finally, the hidden representation is averaged across graph vertices and $p$ fully connected layers are applied sequentially.

\subsection{GCN Fusion}
\subsubsection{Graph Convolutional Layer}
Due to weight sharing between graph vertices, the weights and biases of the GCN have exactly the same structure as those of the MLPs. Hence, the weight-based alignment for GCN layers works exactly the same as for MLPs.

However, the pre-activations of graph convolutional layers are different from the ones in MLPs, namely, they are in the form of graphs. This leaves room for exploiting the graph structure to potentially improve OT fusion performance. During the fusion process we pass the same graphs through both networks and keep track of the matching graph vertices in between networks. This means that when we pass a batch of graphs to the network, each neuron of a layer has a set of graphs associated with its output. The structure of the graphs between neurons and between networks is the same, meaning that there is no need to additionally match graph vertices.
For information on how to fuse other components of GCNs, i.e. biases, BN, and embedding layer see \cref{ssec:bn_intro}.

\subsubsection{Optimal Transport Cost}
\label{sssec:ot_cost}
Solving the OT problem for fusing two GCN layers requires defining the cost of transporting a source point to a target point. Here, a point corresponds to a set of scalar activation graphs. To compute a distance between such sets, we first define a pairwise distance between scalar graphs. Let $(G_i,G_j)$ be a pair of graphs with identical sets of vertices $V$ and edges $E$ and $a_i(u) \in \mathbb{R}$ and $a_j(u) \in \mathbb{R}$ their vertex features, respectively.

We consider three different ways to define a transport cost $c_{ij}$:
(i) {\it Euclidean Feature Distance (EFD)}, where we vectorize both graphs and compute the Euclidean distance between the resulting vectors: $c_{ij}^2 := \lambda \sum_{(u, w) \in E}(a_i(u) - a_j(w))^2$; (ii) {\it Quadratic Energy (QE)}, where we compute the cost as:
        $c_{ij} := \lambda \sum_{(u, w) \in E}(a_i(u) - a_j(w))^2  + (1 - \lambda) \sum_{u \in V}(a_i(u) - a_j(u))^2,$
    with $\lambda \in [0,1]$; 
    (iii) {\it Fused Gromov-Wasserstein (FGW),} where we consider the fused Gromov-Wasserstein distance between graphs $G_i$ and $G_j$ as the pairwise cost between them. This includes solving another OT problem mentioned in \cref{sec:sinkhorn}.

Finally, to compute the cost between two sets of graphs, aligned in terms of sharing the same set of matched vertices and edges, we sum all pairwise costs computed using one of the proposed methods. This provides the optimal transport cost needed for our algorithm.




\section{Experiment Setup}
For individual models' definition as well as their training and evaluation we rely on a benchmarking framework for GNNs, developed by V. P. Dvived et al. in 2018 \cite{dwivedi2020benchmarkgnns} and the ZINC dataset they provide for graph regression tasks.  The dataset consists of 12k molecular graphs (10k-1k-1k:train-test-val), where 28-dimensional node features represent atom types and edges chemical bonds. The quantity to be regressed is a molecular property known as constrained solubility. 

In \cref{sec:results,sec:results-further}, we present the results of five experiments. Unless otherwise stated, we use two GCN models with $g=4$ GC layers with BN, a hidden dimension of 145, and $p=3$ fully connected layers trained until convergence. This resulted in training lengths equal on average to 198 and 106 epochs for GCNs and MLPs respectively. 
\section{Results} \label{sec:results}

\paragraph{Combinations of Different Costs and Optimal Transport Algorithms}
With the first experiment, we determine the best combination of OT costs and algorithms. In doing so, we also investigate whether it is beneficial to use distances considering the underlying graph structure. We first fine-tune the OT unbalancedness constant $\rho=\rho_A=\rho_B$ and the entropy regularization $\epsilon$ (see \cref{ssec:hp-search}) and then report the results for the best configuration. \Cref{tab:results_opt} shows the mean and the standard deviation across five runs using different subsets of samples for the computation of the pre-activations. For EFD and QE, we use a sample size of 100 for the pre-activation computation, and for FGW only 2, because of the significantly larger runtimes it induces. 


\begin{table}[ht]
\begin{center}
\caption{Best results for each pair of fusion type and graph cost. Sample size used for merging was set to 340. Results show the mean and standard deviation of MAE on the ZINC dataset.}
\label{tab:results_opt}
\vskip 8pt
\begin{tabular}{l|cccr}
Fusion Type & EFD & QE & FGW \\
\hline
Emd    & 1.22$\pm$ 0.11& 1.23$\pm$ 0.26& 1.74$\pm$ 0.45\\
Sinkhorn & 1.37$\pm$ 0.31& 1.29$\pm$ 0.23& 1.53$\pm$ 0.26\\
\end{tabular}
\end{center}
\vskip -0.15in
\end{table}

\paragraph{Comparison to Baseline}
In this experiment, we compare the optimal fusion configuration, EMD with EFD, to three baselines: individual models, an ensemble, which averages the predictions of individual models, and vanilla fusion, where the weights of each layer are simply averaged. Unlike in \cref{tab:results_opt}, where we set the sample size for merging to 100, we use a sample size of 340 (see \cref{exp:2} for more details). \Cref{tab:results_comp} compares the MAE of the models and additionally displays the smallest MAEs of our model and the vanilla fusion model after fine-tuning them for 40 epochs.

\begin{table}[ht]
\caption{Results for the best combination of OT fusion algorithm and cost and multiple baselines.}
\label{tab:results_comp}
\vskip 8pt
\begin{center}
\begin{tabular}{l|cccr}
Model & Base & Finetuned \\
\hline
Individual     & [0.42, 0.42]& -\\
Ensemble     & 0.40 & -\\
Emd-EFD    & 1.12$\pm$ 0.07& 0.45$\pm$ 0.02 \\
Vanilla & 2.21 & 0.45\\
\end{tabular}
\end{center}
\vskip -0.15in
\end{table}

\paragraph{Fusing MLPs}
\label{exp:5}

In the last experiment, we compare the fusion of GCNs and MLPs. Both use individual models with the same number of parameters ($\sim$ 100k) trained without BN and we fuse them with the exact OT fusion algorithm and the EFD cost. \Cref{tab:results_mlp} compares the results between GCN and MLP for the individual models, EMD-EFD and vanilla fusion.

\begin{table}[ht]
\caption{Fusion comparison between MLP and GCN.}
\label{tab:results_mlp}
\vskip 8pt
\begin{center}
\begin{tabular}{l|cccr}
Model & MLP & GCN \\
\hline
Individual    & [0.71, 0.70]& [0.48, 0.46] \\
Emd-EFD     & 1.05 $\pm$ 0.02 & 1.05 $\pm$ 0.08 \\
Vanilla    & 1.59 & 1.85  \\
\end{tabular}
\end{center}
\vskip -0.15in
\end{table} 

\section{Discussion}
\vspace{-0.1cm}
\label{sec:discussion}
\Cref{tab:results_opt} shows, that our method significantly exceeds vanilla fusion, implying that OT fusion can be successfully applied to GCNs. However, after fine-tuning, the models achieve the same results, which cannot match the ensemble's or the individual models' MAEs.

The EMD-EFD pairing emerging as the best approach to fusion on GCNs (\cref{tab:results_comp}) may imply that introducing graph structure to the computation of the pairwise graph cost does not help improve fusion. However, it also needs to be mentioned that approaches relying on the FGW graph cost were not entirely fairly compared with the others as far fewer samples could be used for the cost computation. Hence, it would be desirable to find a faster implementation that is also feasible for larger sample sizes.

\Cref{tab:results_mlp}, where we compare GCN and MLP fusion, may imply that GCNs are more difficult to fuse than MLPs. This is supported by the fact that both the vanilla fusion model and our EMD-EFD model are closer to the performance of the individual models for MLPs. This potentially relates to the difference in optimization landscapes for the two different architecture types, as they have a large impact on the geometry of the activation space. 

\section{Conclusion}
\vspace{-0.1cm}
\paragraph{Summary.} In this paper, we presented an approach for successfully fusing GCNs through layer-wise weight alignment with OT. Our results suggest that fusing GCNs is more challenging than fusing MLPs and that incorporating the graph structure does not improve the fusion process.

\paragraph{Limitations and future work.}  Our experiments incorporate the graph structure using only two approaches (see QE and FGW in \cref{sssec:ot_cost}). Further experimentation is necessary to fully determine whether the graph structure contributes to the OT fusion process.

In future work, our approach could be tested on additional graph datasets for vertex, edge, and graph-type prediction. It could also be extended to merge more than two GCNs and evaluated in the context of one-shot skill transfer~\cite{singh2023model}.



\section*{Acknowledgements}
We want to thank Sidak Pal Singh (Data Analytics Lab, Department of Computer Science, ETH Zürich) for our fruitful discussions about the project. Michael Vollenweider would like to acknowledge the financial support from the Seminar of Statistics, ETH Zurich. Michael Vollenweider and Elisa Hoskovec would like to acknowledge the financial support from the Data Analytics Lab, Department of Computer Science, ETH Zürich. 

\bibliography{./iclr2025_conference}
\bibliographystyle{iclr2025_conference}

\appendix
\section{OT Solvers}

\Cref{sec:sinkhorn} describes the different algorithms used for OT. In our project, we use the implementation by \cite{flamary2021pot} for the EMD algorithm, and the fixed point iteration solver implemented by \cite{cuturi2022optimal} for the first version of the Sinkhorn algorithm.


\section{Other GCN components}
\label{ssec:bn_intro}
{\bf Bias:} Each term is connected only to a single output of its layer, hence we simply pre-multiply it by the transpose of the same TM as for the layer weights. 

{\bf Batch Normalisation:} For each input, BN learns two weights and estimates input mean and variance. Both the learnable parameters and the estimated statistics need to be aligned for fusion. No aggregation operation would result in any interconnections between different inputs, hence there is no need to compute a TM for BN layers. The parameters and statistics are aligned using the TM of the previous layer and the same TM is propagated further. For the networks with BN layers, we consider two approaches: for a layer $L$ followed by BN we experiment with taking either the pre-activations of layer $L$ or the pre-activations of the BN following layer $L$ (see \cref{ssec:bn}).

{\bf Embedding Layer:} For a one-hot encoded input (as the atom type in the ZINC dataset) it is just an MLP with no bias term. Hence, we align embedding layers the same way as MLPs.

\section{Further Results}
\label{sec:results-further}
This section presents further experiments and their results.

\subsection{Hyperparameter search}
\label{ssec:hp-search}
For Sinkhorn, all three costs achieve their best MAE with \(\rho = 1\), but differ in their optimal regularization parameter \(\epsilon\): For QE and FGW \(\epsilon = 5\times10^{-5}\), and for EFD \(\epsilon = 5\times10^{-4}\) are optimal. Moreover, for QE, \(\lambda = 0.2\) performs best for both OT algorithms.

\subsection{Optimal Number of Samples for Best Combo} \label{exp:2}
\begin{figure}[h]
    \vspace{-0.1in}
    \centering
    \includegraphics[width=0.8\linewidth]{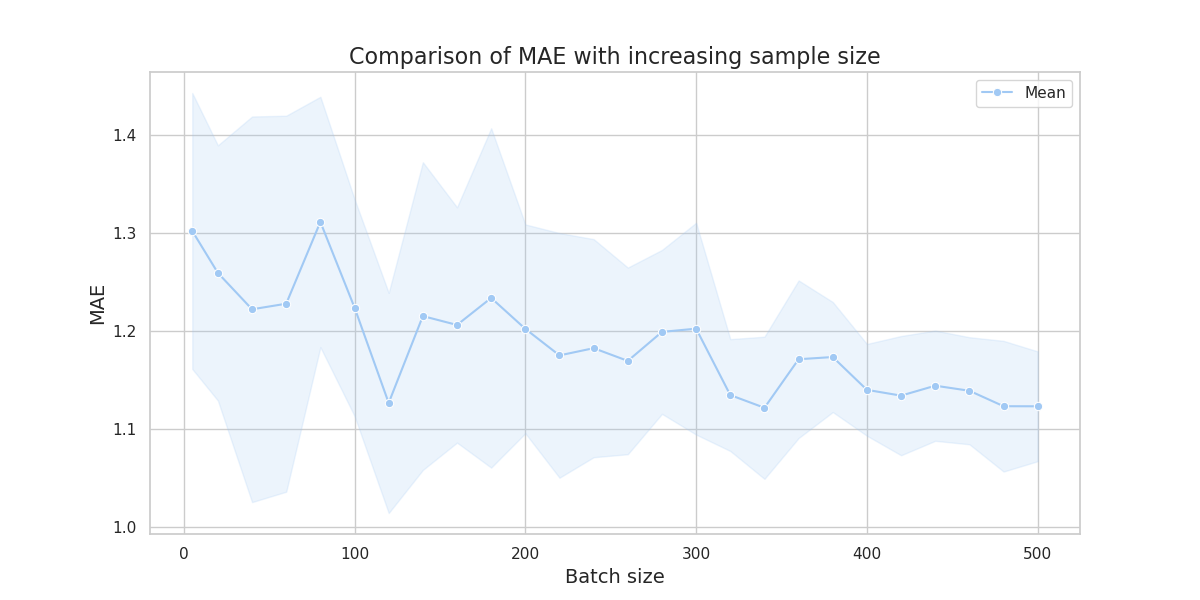}
    \caption{Influence of the sample size of pre-activations on MAE. The line represents the mean, the shading the standard deviation.}
    \label{fig:batchsize-plot}
    \vspace{-0.1in}
\end{figure}

This experiment determines the optimal pre-activation sample size (referred to as batch size in \cref{fig:batchsize-plot}) for the best configuration from the first experiment (see \cref{tab:results_opt}). \Cref{fig:batchsize-plot} displays the influence of the increasing sample size on the MAE.

MAE decreases strongly at first and reaches a near-optimum at a sample size of 120. Sample sizes of 340, 480, and 500 all reach the optimal mean MAE of 1.12.

The results imply that the fusion performance generally increases with the larger size of the sample used to calculate the pairwise graph costs. This is not surprising, as using larger sample sizes corresponds to giving the fusion algorithm more information about the geometry of the activation landscape. 

\subsection{Influence of Batch Normalization}
\label{ssec:bn}

As mentioned in \cref{ssec:bn_intro} for networks with BN we consider using pre-activations from before and after BN. Using EMD with EFD this results in a mean MAE of 2.04$\pm$ 0.41 and 1.12$\pm$ 0.07, respectively. We conclude that taking the pre-activations after BN significantly improves the results for our optimal configuration.





\section{Deep Graph Mating}
\label{app:grama}
Recent work by \cite{jing2024deep} introduces a training-free approach to fusing graph neural networks (GNNs) by aligning neuron permutations before parameter interpolation. Their method formulates the problem as a linear assignment problem (LAP), optimizing permutation matrices in a topology-aware manner through their dual-message coordination and calibration (DuMCC) framework.

Our approach, in contrast, formulates model fusion as an optimal transport (OT) problem, aligning weights based on transport costs rather than discrete assignments. While GRAMA incorporates graph structure into the alignment process through message coordination, our method explores different cost functions to determine their effect on fusion quality.

Both methods contribute to the broader goal of training-free model fusion, and future work could investigate potential connections between OT-based and LAP-based alignment strategies.

\end{document}